\def\year{2022}\relax
\documentclass[letterpaper]{article} 
\usepackage{aaai22}  
\usepackage{times}  
\usepackage{helvet}  
\usepackage{courier}  
\usepackage[hyphens]{url}  
\usepackage{graphicx} 
\usepackage{natbib}  
\usepackage{caption} 
\DeclareCaptionStyle{ruled}{labelfont=normalfont,labelsep=colon,strut=off} 
\usepackage{algorithm}
\usepackage{algorithmic}
\usepackage{amsmath}
\usepackage{amsfonts}
\usepackage{xcolor}
\usepackage{multirow}
\usepackage{array}

\usepackage{newfloat}
\usepackage{listings}
\floatstyle{ruled}
\newfloat{listing}{tb}{lst}{}
\floatname{listing}{Listing}
\title{Multi-Source Video Domain Adaptation with\\ Temporal Attentive Moment Alignment Network}

\author{
    Yuecong Xu,\equalcontrib\textsuperscript{\rm 2}
    Jianfei Yang,\equalcontrib\textsuperscript{\rm 1}
    Haozhi Cao,\textsuperscript{\rm 1}\\
    Keyu Wu,\textsuperscript{\rm 2}
    Min Wu, \textsuperscript{\rm 2}
    Rui Zhao, \textsuperscript{\rm 3}
    Zhenghua Chen \textsuperscript{\rm 2}
}
\affiliations{
    \textsuperscript{\rm 1}School of Electrical and Electronic Engineering, Nanyang Technological University, Singapore,\\
    \textsuperscript{\rm 2}Institute for Infocomm Research, A*STAR, Singapore,
    \textsuperscript{\rm 3}Pluang,\\
    \{xuyu0014, yang0478, haozhi001\}@e.ntu.edu.sg,
    \{wu\textunderscore keyu, wumin\}@i2r.a-star.edu.sg,\\
    zhao.rui@pluang.com,
    chen0832@e.ntu.edu.sg
}

\begin{document}
\hbadness=1000000000
\vbadness=1000000000
\hfuzz=10pt

\setlength{\abovedisplayskip}{2pt}
\setlength{\belowdisplayskip}{2pt}
\setlength{\textfloatsep}{10pt plus 1.0pt minus 2.0pt}
\setlength{\floatsep}{6pt plus 1.0pt minus 1.0pt}
\setlength{\intextsep}{6pt plus 1.0pt minus 1.0pt}

\maketitle

\begin{abstract}
    Multi-Source Domain Adaptation (MSDA) is a more practical domain adaptation scenario in real-world scenarios. It relaxes the assumption in conventional Unsupervised Domain Adaptation (UDA) that source data are sampled from a single domain and match a uniform data distribution. MSDA is more difficult due to the existence of different domain shifts between distinct domain pairs. When considering videos, the negative transfer would be provoked by spatial-temporal features and can be formulated into a more challenging Multi-Source Video Domain Adaptation (MSVDA) problem. In this paper, we address the MSVDA problem by proposing a novel Temporal Attentive Moment Alignment Network (TAMAN) which aims for effective feature transfer by dynamically aligning both spatial and temporal feature moments. TAMAN further constructs robust global temporal features by attending to dominant domain-invariant local temporal features with high local classification confidence and low disparity between global and local feature discrepancies. To facilitate future research on the MSVDA problem, we introduce comprehensive benchmarks, covering extensive MSVDA scenarios. Empirical results demonstrate a superior performance of the proposed TAMAN across multiple MSVDA benchmarks.

\end{abstract}

\section{Introduction}
\label{section:intro}

Video-based tasks (e.g., action recognition) have long been researched considering their wide applications. Among the different methods proposed, neural networks have made remarkable advances in these tasks due to the large-scale labeled datasets for training and testing. Yet sufficient labeled training videos may not be readily available in real-world scenarios owing to the high cost of video data annotation. Subsequently, various \textit{Unsupervised Domain Adaptation} (UDA) and \textit{Video-based Unsupervised Domain Adaptation} (VUDA) methods have been introduced to transfer knowledge from a labeled source domain to an unlabeled target domain by reducing discrepancies between source and target domain.

\begin{figure}[t]
\begin{center}
   \includegraphics[width=1.\linewidth]{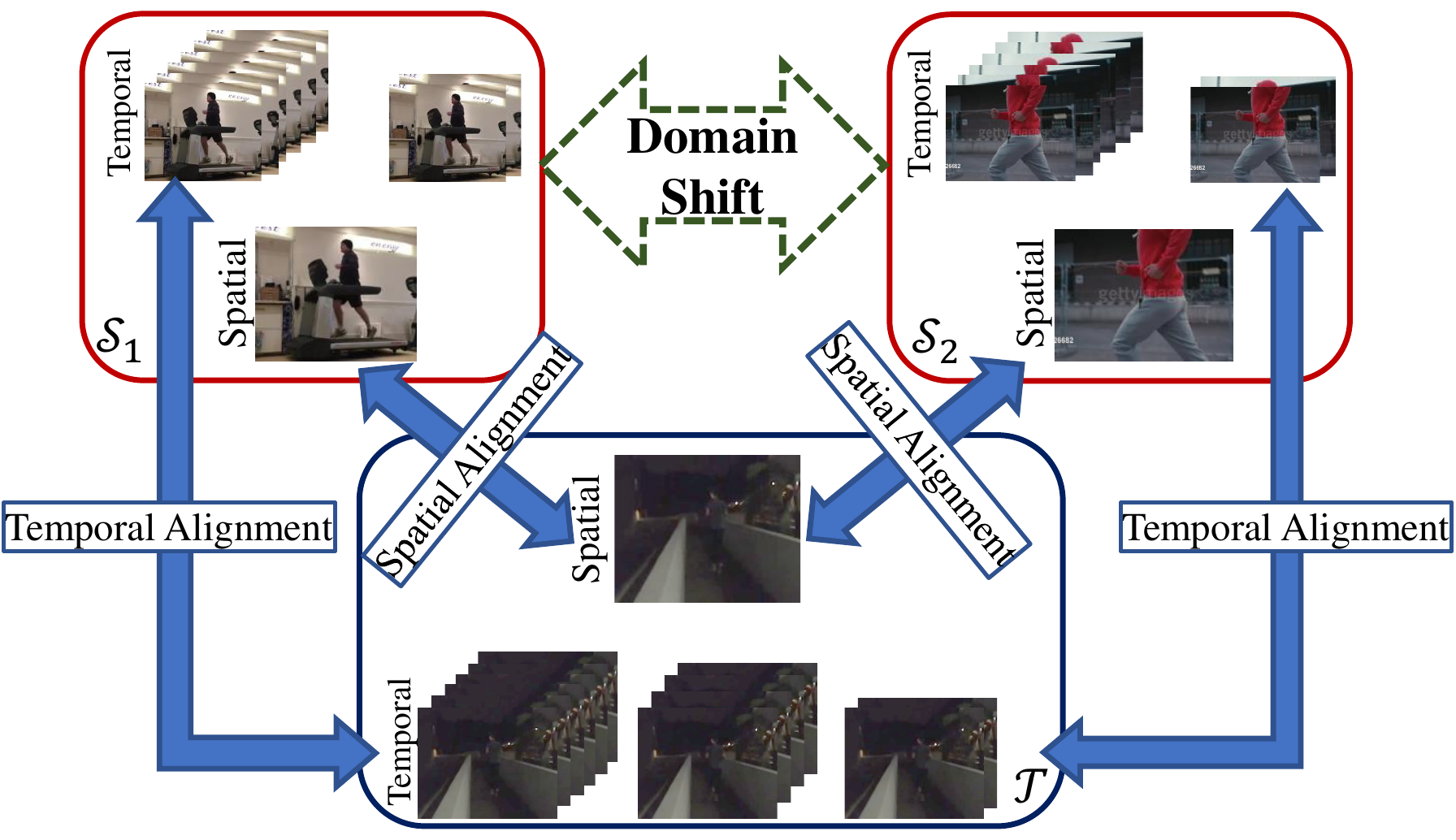}
\end{center}
   \caption{MSVDA is more generic compared to VUDA where source data come from multiple domains, with different data distributions. MSVDA is more challenging due to the negative transfer caused by domain shifts among source domains (depicted as dashed arrow) and the need to jointly align the target domain $\mathcal{T}$ and the different source domains $\mathcal{S}_{1}$ and $\mathcal{S}_{2}$. Such negative transfer could be provoked by both spatial and temporal features in MSVDA.}
\label{figure:1-1-intro}
\end{figure}

Though current UDA and VUDA methods~\cite{zhang2019bridging,xu2021aligning} enable the transfer of knowledge across domains, they normally assume that the training source data are sampled from a single domain and match a uniform data distribution. Such assumption may not hold in real-world applications. In practice, with the availability of different large-scale labeled public datasets, training source data are more likely to be collected from multiple datasets. This scenario is defined as \textit{Multi-Source Domain Adaptation} (MSDA) that relaxes the constraint of identical source data distribution by assuming that source data are sampled from multiple domains corresponding to different data distributions. The MSDA problem is more difficult owing to the existence of different levels of domain shifts among source domains and between different source-target domain pairs, which adversely affects the alignment of target data, resulting in \textit{negative transfer}.

In contrast with images that only contain spatial features, temporal features are key components in video representations that are not contained in images. The presence of the additional features engenders a novel \textit{Multi-Source Video Domain Adaptation} (MSVDA) problem, which aims to transfer networks trained with data from multiple source domains to the target domain. MSVDA empowers models trained in a collection of large-scale video datasets such as UCF101~\cite{soomro2012ucf101} and HMDB51~\cite{kuehne2011hmdb} to be employed directly to smaller-scale datasets such as ARID~\cite{xu2021arid} without label supervision. For MSVDA, negative transfer would be triggered if we directly reduce the divergence between multiple domain pairs regardless of inconsistent domain shifts caused by distinct spatial and temporal feature distributions, as presented in Figure~\ref{figure:1-1-intro}.

To tackle negative transfer in MSVDA, we argue that the additional temporal features should be utilized from two perspective: firstly, effective global temporal features should be constructed with attention to dominant local temporal features with higher local classification confidence, alleviating the probability of provoking negative transfer by temporal features; secondly, the temporal features should contribute towards the overall feature alignment process together with the spatial features, eliminating the possibility of misalignment of spatial features. To this end, we propose a novel \textbf{Temporal Attentive Moment Alignment Network (TAMAN)} to address the challenges in MSVDA uniformly. TAMAN first constructs robust global temporal features by attentive combination of local temporal features which represent the different characteristics of the overall motion. The attention strategies depend on both the local temporal feature classification confidence, as well as the disparity between global and local feature discrepancies. Meanwhile, TAMAN aligns spatial-temporal features jointly by aligning the moments of both spatial and temporal features across all domain pairs, mitigating possible negative transfer caused by misalignment of spatial features.

To aid MSVDA research, we propose two sets of comprehensive benchmarks, fully utilizing both widely used public datasets in action recognition and a more recent dataset built with dark videos. The proposed benchmarks are: (i) \textit{Daily-DA}, constructed with the ARID~\cite{xu2021arid}, HMDB51~\cite{kuehne2011hmdb}, Kinetics~\cite{kay2017kinetics}, and Moments-in-Time~\cite{monfort2019moments} datasets; and (ii) \textit{Sports-DA}, constructed with the UCF101~\cite{soomro2012ucf101}, Sports-1M~\cite{karpathy2014large}, and Kinetics datasets. The proposed benchmarks cover extensive MSVDA scenarios with distinct domain shifts across included domains.

In summary, our contributions are threefold. Firstly, we formulate a novel practical and challenging \textit{Multi-Source Video Domain Adaptation} (MSVDA) problem. To the best of our knowledge, this is the first research that investigates multi-source domain adaptation in the video classification field, especially action recognition. Secondly, we analyze the challenges of the MSVDA problem and propose TAMAN to address the challenges. TAMAN learns robust global temporal features with local temporal attention strategies, while utilizing moments of both spatial and temporal features for feature alignment across domain pairs jointly. Finally, we introduce two sets of MSVDA benchmarks and exhibit the capability of TAMAN, achieving superior performances across all the proposed MSVDA benchmarks.

\section{Related Work}
\label{section:related}

\textbf{Unsupervised Domain Adaptation (UDA).} 
Current UDA methods aim to distill shared knowledge across domains with the labeled source domain and unlabeled target domain, thus improving the transferability of models. In general, these methods could be divided into three categories: a) reconstruction-based methods~\cite{ghifary2016deep,jhuo2012robust,aljundi2016lightweight}, where domain-invariant features are obtained by encoders trained under data-reconstruction schemas, typically formulated as encoder-decoder networks; b) adversarial-based methods~\cite{ganin2015unsupervised,tzeng2017adversarial,zou2019consensus}, which are inspired by the success of GAN~\cite{goodfellow2014generative}, are designed with additional domain discriminators that are trained jointly with feature generators in an adversarial manner~\cite{huang2011adversarial}, minimizing adversarial losses~\cite{ganin2015unsupervised}; and c) discrepancy-based methods~\cite{long2015learning,saito2018maximum,zhang2019bridging}, which alleviate domain shifts across source-target domain pairs by employing various metric learning schemas, including MMD~\cite{long2015learning}, CORAL~\cite{sun2016return} and KL-divergence~\cite{zhuang2015supervised}. Discrepancy-based methods do not require additional network structures (e.g., domain classifiers), thus are more stable and easy to train. More recently, with the wide applications of videos in various fields, there has been increasing research for Video-based Unsupervised Domain Adaptation (VUDA). The success of obtaining domain-invariant features with the above UDA methods extends to VUDA, with multiple VUDA methods proposed for tasks such as action recognition~\cite{chen2019temporal,pan2020adversarial,xu2021aligning} and action segmentation~\cite{chen2020action}. 

\textbf{Multi-Source Domain Adaptation (MSDA).}
Though UDA and VUDA methods have made outstanding progress, current approaches generally assume that the training source data are sampled from a single domain and follow a uniform data distribution. A more general and practical scenario that relaxes this assumption is denoted as Multi-Source Domain Adaptation (MSDA)~\cite{mansour2008domain,hoffman2012discovering}, which enables models to transfer knowledge from multiple sources. Earlier MSDA methods rely on either hand-crafted feature representations~\cite{sun2011two,duan2012exploiting} or pre-trained classifiers~\cite{xu2012multi,sun2013bayesian}. These works demonstrate the applications of MSDA in fields such as image classification~\cite{mansour2008domain,sun2011two} and multimedia classification~\cite{duan2012exploiting,chattopadhyay2012multisource}. More recently, with the advances in deep neural networks, various end-to-end MSDA methods have been proposed. Among these, MDAN~\cite{zhao2018adversarial} aligns the target domain to source domains globally with adversarial learning, applying a domain discriminator for each source-target domain pair and a single task classifier. DCTN~\cite{xu2018deep} improves on MDAN by deploying a separate task classifier for each source domain, with the final result being a weighted combination of the output predictions. Further, MDDA~\cite{zhao2020multi} introduces a source distillation mechanism for fine-tuning both the feature extractor and the task classifier while CMSS~\cite{yang2020curriculum} introduces a dynamic curriculum that updates the error rate of domain discriminators constantly. Meanwhile, M3SDA~\cite{peng2019moment} utilizes a moment matching component for transferring knowledge.

Despite the notable progress made in MSDA with its various applications, current MSDA approaches are mostly built for image-based MSDA, with both the source and target domains being image data. Meanwhile, MSVDA, which focuses on video-based knowledge transfer, has not been dealt with. MSVDA is more challenging due to the possibility that negative transfer could be triggered by temporal features, which do not exist in images. We propose to tackle MSVDA with a novel method that constructs robust global temporal features with local temporal attention strategies while utilizing moments of both spatial and temporal features for effective feature alignment.

\section{Proposed Method}
\label{section:method}

In the scenario of \textit{Multi-Source Video Domain Adaptation} (MSVDA), we are given a collection of $M$ source domains denoted as $\mathcal{S}=\{\mathcal{S}_{1}, \mathcal{S}_{2}, ... , \mathcal{S}_{M}\}$, with domain $\mathcal{S}_{m}=\{(V_{i\mathcal{S}_{m}},y_{i\mathcal{S}_{m}})\}^{n_{\mathcal{S}_{m}}}_{i=1}$ containing $n_{\mathcal{S}_{m}}$ i.i.d.\ labeled videos associated with $K$ classes and characterized by a probability distribution of $p_{\mathcal{S}_{m}}$. A target domain $\mathcal{T}=\{V_{i\mathcal{T}}\}^{n_{\mathcal{T}}}_{i=1}$ with $n_{\mathcal{T}}$ i.i.d.\ unlabeled videos characterized by a probability distribution of $p_{\mathcal{T}}$ is accessed. We assume that the unlabeled target domain videos share the same $K$ classes with the labeled source domain videos. To tackle the MSVDA problem, our goal is to build a robust network capable of learning transferable features across the multiple video source domains and the video target domain, while minimizing the target classification risk. 

In contrast with conventional VUDA, MSVDA is more challenging owing to the existence of both domain shifts between the different source-target domain pairs and among the different source domains. Moreover, while there are some existing MSDA approaches that tackle the negative effect brought by the extra domain shifts, these approaches are mostly built for image-based MSDA problems with the source data being image data only. Negative transfer in these problems could only be provoked by domain shift for spatial features. Meanwhile, videos contain temporal features which represent the motion information, thus negative transfer could be further triggered by different domain shifts w.r.t. temporal features. Since only spatial features are relevant for image-based MSDA problems, current MSDA approaches may not have the capability for constructing robust global temporal features which further results in their insensitivity towards temporal feature misalignment. Therefore, a novel Temporal Attentive Moment Alignment Network (TAMAN) is introduced to transfer from multiple video source domains while alleviating negative transfer with full usage of effective temporal features built in an attentive manner. We start with a brief review of discrepancy-based MSDA approaches utilizing moment alignment, proceeded by a detailed description of TAMAN.

\begin{figure*}[t]
\begin{center}
   \includegraphics[width=.9\linewidth]{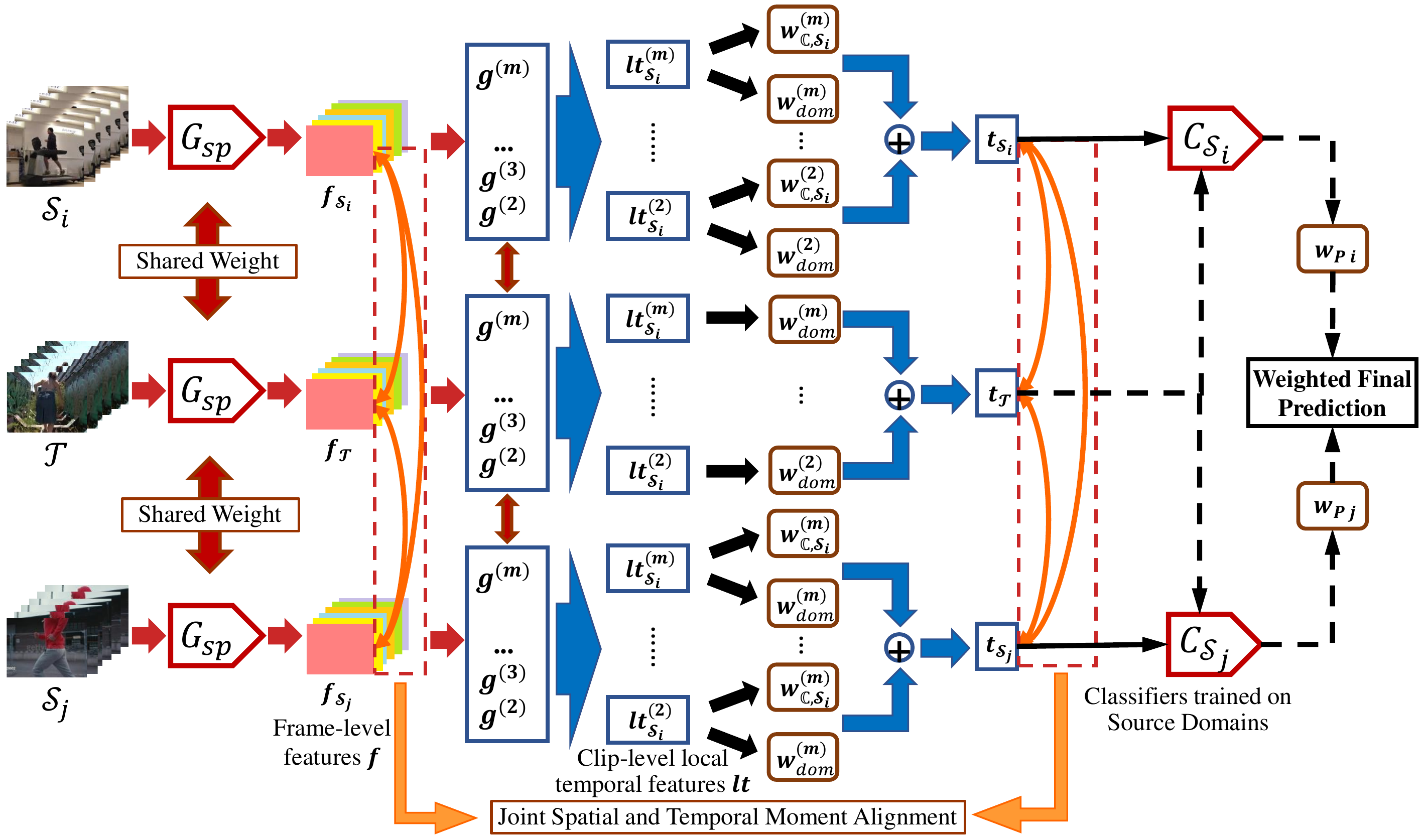}
\end{center}
    \caption{Architecture of the proposed TAMAN. To perform MSVDA effectively, the robust overall feature $\mathbf{t}$ is constructed by attentive aggregation of clip-level local temporal features $\mathbf{lt}^{(r)}$, obtained from time-ordered frame-level spatial features. The weights of the local temporal features include both the \textit{local confidence weight} $w_{\mathbb{C}}^{(r)}$ and the \textit{dominance weight} $w_{dom}^{(r)}$. Moment alignment is performed jointly across the spatial features and global temporal features. Dashed arrows indicate how target data is propagated during testing where the final prediction is obtained from a weighted ensemble schema. 
    }
\label{figure:3-2-taman}
\end{figure*}

\subsection{Discrepancy-based MSDA with Moment Alignment}
\label{section:method:mada}
The goal of conventional UDA and its variants are to align the data distributions of source and target domains. Discrepancy-based approaches are widely used thanks to their ability to alleviate domain shift through metric learning schemas without additional network components. Given the feature extractor $G_{f}$, the source classifier $C_{\mathcal{S}}$, with $\mathbf{x}_{\mathcal{S}}$ and $\mathbf{x}_{\mathcal{T}}$ being collections of $n_\mathcal{S}$ source domain samples and $n_\mathcal{T}$ target domain samples, the overall objective of discrepancy-based DA methods are generally formulated as:
\begin{equation}
\label{eqn:method:dis-da}
    \begin{aligned}
    \mathcal{L}
    &= \frac{1}{n_{\mathcal{S}}}\sum\limits_{x\in \mathbf{x}_{\mathcal{S}}}
    L_{\mathcal{S}}(C_{\mathcal{S}}(G_{f}(x)), y)\\ 
    &+ \lambda_{d}\:d\,(G_{f}(\mathbf{x}_{\mathcal{S}}), G_{f}(\mathbf{x}_{\mathcal{T}})),
    \end{aligned}
\end{equation}
where $L_{\mathcal{S}}$ stands for the source classification loss with $y$ being the ground truth label of input $x$ from the source domain, while $\lambda_{d}$ is the trade-off parameter for the cross-domain discrepancy $d$. While various forms of  discrepancies have been proposed, a major line of which are moment-based. Minimizing such discrepancies could therefore be viewed as moment alignment schemas. Typical examples include MMD~\cite{long2015learning}, which matches the first moments of distributions, and CORAL~\cite{sun2016return}, which matches the second moments of distributions.

While these moment alignment methods could align data distributions under conventional UDA settings, their performances degrade substantially when applying to MSDA tasks, owing to negative transfer caused by the domain shifts between the different source-target domain pairs and within the different source domains. As proven in~\cite{peng2019moment}, the upper bound of the target classification risk relates closely to the pairwise cross-moment discrepancy between the target domain and each source domain, denoted as $d_{CM}(\mathcal{S},\mathcal{T})$, which can be formulated as:
\begin{equation}
\label{eqn:method:ms-dcm}
    d_{CM}(\mathcal{S},\mathcal{T})
    =   \sum\limits_{j=1}^{M}\lambda_{\mathcal{S}_{j}}
        \sum\limits_{k}d_{CM^{k}}(\mathcal{S}_{j},\mathcal{T}),
\end{equation}
where $d_{CM^{k}}(\mathcal{S}_{j},\mathcal{T})$ is the $k$-th moment discrepancy between the $j$-th source domain and the target domain. Since $d_{CM^{k}}(\cdot, \cdot)$ is a metric, it follows the triangle inequality, formulated as:
\begin{equation}
\label{eqn:method:ms-dcm-lb}
    d_{CM^{k}}(\mathcal{S}_{i},\mathcal{S}_{j})
    \leq d_{CM^{k}}(\mathcal{S}_{i},\mathcal{T}) + d_{CM^{k}}(\mathcal{S}_{j},\mathcal{T}).
\end{equation}
This implies that the cross-moment discrepancy between domains $\mathcal{S}_{j},\mathcal{T}$ is lower bounded by the pairwise discrepancies between source domains. Combining Equations \ref{eqn:method:ms-dcm} and \ref{eqn:method:ms-dcm-lb}, a moment distance is introduced as:
\begin{equation}
\label{eqn:method:init-m3sda}
\resizebox{.9\linewidth}{!}{$
    \begin{aligned}
    d_{CM}(\mathcal{S},\mathcal{T})
    &= \sum\limits_{k}(\frac{1}{M}\sum\limits_{i=1}^{M}
\|\mathbf{E}\,({\mathbf{x}_{\mathcal{S}_{i}}}^{k}) - \mathbf{E}\,({\mathbf{x}_{\mathcal{T}}}^{k})\|_{2} \\
    &+ \binom{M}{2}^{-1}\sum\limits_{i,j \in [1, M]}
\|\mathbf{E}\,({\mathbf{x}_{\mathcal{S}_{i}}}^{k}) - \mathbf{E}\,({\mathbf{x}_{\mathcal{S}_{j}}}^{k})\|_{2}
            ).
    \end{aligned}
$}
\end{equation}
The moment distance $d_{CM}$ enables effective multi-source domain adaptation through minimizing both source-target domain discrepancies and discrepancies between different source domains. The overall objective function of MSDA is finally formulated as suggested in~\cite{peng2019moment}:
\begin{equation}
\label{eqn:method:obj-m3sda}
\resizebox{.9\linewidth}{!}{$
    \mathcal{L}_{ms}
    = \sum\limits_{j=1}^{M}\frac{1}{n_{\mathcal{S}_{j}}}\sum\limits_{x\in \mathbf{x}_{\mathcal{S}_{j}}}
    L_{\mathcal{S}_{j}}(C_{\mathcal{S}_{j}}(G_{f}(x)), y)
    + \lambda_{d}\:d_{CM}(\mathcal{S},\mathcal{T}),
    $}
\end{equation}
where $L_{\mathcal{S}_{j}}$ stands for the source classification loss for the $j$-th source classifier with $y$ being the ground truth of input $x$ from the $\mathcal{S}_{j}$, while $\lambda_{d}$ is the trade-off parameter.

\subsection{Temporal Attentive Moment Alignment Network}
\label{section:method:taman}
To achieve MSVDA, an intuitive approach would be to apply the moment matching to video data directly by integrating videos into Equation~\ref{eqn:method:obj-m3sda}, i.e., $x=V$. Meanwhile, the original image feature extractor could simply be substituted with a video feature extractor.

Nevertheless, despite the simplicity of the moment alignment, empirical results suggest that such method is insufficient to deal with negative transfer in MSVDA well, leading to an inferior adaptation result. We can expect such inferior results, since the video representations obtained through common feature extractors (i.e., convolutional neural network (CNN)-based extractors) focus primarily on spatial features. In contrast, temporal features are normally obtained vaguely with a simple pooling process across the temporal dimension, leading to tremendous distribution shifts. Without explicit temporal features, domain adaptation through moment alignment may only be performed on spatial features. The negative transfer provoked by the temporal features could not be dealt with by simply reducing moment distance.

In light of the above shortcomings, we introduce a novel \textbf{Temporal Attentive Moment Alignment Network (TAMAN)} to perform multi-source adaptation by exploiting both spatial and temporal features, with its structure shown in Figure~\ref{figure:3-2-taman}. To enable temporal features for moment alignment, a key prior is to obtain temporal features explicitly. Compared to conventional CNN-based extractors (e.g., 3D-ResNet~\cite{hara2017learning}) whose temporal features are obtained via temporal pooling, Temporal Relation Network (TRN)~\cite{zhou2018temporal} is preferred. This is thanks to its ability to extract temporal features through reasoning over the correlations between spatial representations, which coincides with the human approach on recognizing actions. With the frame-level spatial features obtained from the shared spatial feature extractor $G_{sp}$, the $v$-th input video from domain $\mathcal{S}_{m}$ with $h$ frames is expressed as $V_{v\,\mathcal{S}_{m}}=\{f_{v\,\mathcal{S}_{m}}^{(1)},f_{v\,\mathcal{S}_{m}}^{(2)}, ... ,f_{v\,\mathcal{S}_{m}}^{(h)}\}$. Here $f_{v\,\mathcal{S}_{m}}^{(i)}$ is the $i$-th frame-level spatial feature of the $v$-th video from domain $\mathcal{S}_{m}$. For clarity, the subscript $v$ is omitted in subsequent equations. TRN constructs the global temporal features of $V_{\,\mathcal{S}_{m}}$ denoted by $\mathbf{t}_{\,\mathcal{S}_{m}}$ by an aggregation of multiple clip-level local temporal features, each of which are built from $r$ temporal-ordered frames with $r\in [2,h]$. A clip-level local temporal feature is defined as:
\begin{equation}
\label{eqn:method:lt}
    lt_{\,\mathcal{S}_{m}}^{(r)} = \sum\nolimits_{z} g_{lt}^{(r)}((V_{v\,\mathcal{S}_{m}}^{(r)})_{z}).
\end{equation}
Here $(V_{\,\mathcal{S}_{m}}^{(r)})_{z}=\{f_{\,\mathcal{S}_{m}}^{(a)}, f_{\,\mathcal{S}_{m}}^{(b)}, ..., \}_{z}$ represents the $z$-th clip that contains $r$ temporal-ordered frames, with frame indices $a$ and $b$. Note that $b>a$ but $a$ and $b$ can be unconsecutive. The local temporal feature $lt_{\,\mathcal{S}_{m}}^{(r)}$ is computed by integrating the collection of temporal-ordered frame-level spatial features through a integration function $g_{lt}^{(r)}$. The integration function is implemented as a Multi-layer Perceptron (MLP).

The global temporal features could be obtained by simple aggregation strategies applied to all local temporal features (e.g., an average operation). However, the contribution of each local temporal feature is empirically not equal, which motivates us to develop a \textit{local attention} mechanism with two attention strategies to attend to dominant domain-invariant local temporal features. Firstly, inspired by findings in~\cite{chen2019temporal}, we enable TAMAN to focus on more transferable local temporal features. To this end, TAMAN learns global temporal features that attend to more transferable local temporal features, which correspond to higher local class prediction confidence. Specifically, the prediction of each local temporal feature is first obtained by applying the classifier of domain $\mathcal{S}_{m}$ to feature $lt_{\,\mathcal{S}_{m}}^{(r)}$, denoted as $\hat{y}_{lt,\,\mathcal{S}_{m}}^{r} = C_{\mathcal{S}_{m}}(lt_{\,\mathcal{S}_{m}}^{(r)})$, and indicates the probability of the local temporal feature classified as each video class. Suppose that there are a total of $\mathcal{C}$ video classes, the confidence of prediction $\hat{y}_{lt,\,\mathcal{S}_{m}}^{r}$ is defined as the additive inverse of its entropy computed over all the probabilities, formulated as:
\begin{equation}
\label{eqn:method:lt-confidence}
    \mathbb{C}(\hat{y}_{lt,\,\mathcal{S}_{m}}^{(r)}) = \sum\limits_{c=1}^{\mathcal{C}} \hat{y}_{lt,\,\mathcal{S}_{m},\,c}^{(r)}log(\hat{y}_{lt,\,\mathcal{S}_{m},\,c}^{(r)}),
\end{equation}
where $\hat{y}_{lt,\,\mathcal{S}_{m},\,c}^{(r)}$ corresponds to the prediction of the $c$-th class. The \textit{local confidence weight} corresponding to the local temporal feature $lt_{\,\mathcal{S}_{m}}^{(r)}$ is generated by adding a residual connection for more stable optimization, and a $tanh$ function for constraining the weights within the range of $[0,1]$. The formulation of the \textit{local confidence weight} is thus:
\begin{equation}
\label{eqn:method:conf-w}
    w_{\mathbb{C},\,\mathcal{S}_{m}}^{(r)} = \tanh(1 + \mathbb{C}(\hat{y}_{lt,\,\mathcal{S}_{m}}^{(r)})).
\end{equation}

Secondly, inspired by temporal action localization~\cite{shou2016temporal} and action detection tasks~\cite{zhao2017temporal}, it is thought that most actions would be observed in a local temporal range, therefore effective global temporal features should be constructed by focusing on the dominant local temporal feature, discarding the ineffective clips that may lead to domain shifts. Due to the fact that target videos are unlabeled, it is impossible to obtain the prediction accuracies of each local temporal feature. Instead, the \textit{dominance weight} is generated by the disparity between the global and local temporal feature discrepancies. Formally, the raw global temporal feature from the source domain $\mathcal{S}_{m}$ and the target domain $\mathcal{T}$, denoted as $\hat{\mathbf{t}}_{\mathcal{S}_{m}}$ and $\hat{\mathbf{t}}_{\mathcal{T}}$ are obtained by a simple additive aggregation of the clip-level local temporal features, i.e., $\hat{\mathbf{t}}_{\mathcal{S}_{m}} = \sum\limits_{r}\,lt_{\,\mathcal{S}_{m}}^{(r)}$ and $\hat{\mathbf{t}}_{\mathcal{T}} = \sum\limits_{r}\,lt_{\,\mathcal{T}}^{(r)}$. The feature discrepancy is defined based on the cross-moment discrepancy in Equation~\ref{eqn:method:init-m3sda}, where the moment-based local temporal discrepancy $d_{lt}^{(r)}$ is formulated as:
\begin{equation}
\label{eqn:method:dis-feat}
\resizebox{.9\linewidth}{!}{$
    \begin{aligned}
    d_{lt}^{(r)}(\mathcal{S},\mathcal{T})
    &= \sum\limits_{k}(\frac{1}{M}\sum\limits_{i=1}^{M}
\|\mathbf{E}\,(({lt_{\mathcal{S}_{i}}^{(r)}})^{k}) - \mathbf{E}\,(({lt_{\mathcal{T}}^{(r)}})^{k})\|_{2} \\
    &+ \binom{M}{2}^{-1}\sum\limits_{i,j \in [1, M]}
\|\mathbf{E}\,(({lt_{\mathcal{S}_{i}}^{(r)}})^{k}) - \mathbf{E}\,(({lt_{\mathcal{S}_{j}}^{(r)}})^{k})\|_{2}
            ).
    \end{aligned}
$}
\end{equation}
The global temporal discrepancy $d_{\hat{\mathbf{t}}}$ is defined similarly. The \textit{dominance weight} is subsequently generated by the disparity between $d_{\hat{\mathbf{t}}}$ and $d_{lt}^{(r)}$, computed as $d_{d}^{(r)} = |d_{\hat{\mathbf{t}}} - d_{lt}^{(r)}|$. The \textit{dominance weight} $w_{dom}^{(r)}$ is therefore formulated as:
\begin{equation}
\label{eqn:method:dom-w}
    w_{dom}^{(r)} = \mathbf{e}^{d_{d}^{(r)}}/\sum\nolimits_{r}\mathbf{e}^{d_{d}^{(r)}}.
\end{equation}

Finally, the global temporal feature is an attentive aggregation of all local temporal features, with the \textit{local attention weight} $w_{\mathcal{S}_{m}}^{(r)}$ being the multiplication of the \textit{local confidence weight} and the \textit{dominance weight}, i.e., $w_{\mathcal{S}_{m}}^{(r)} = w_{\mathbb{C},\,\mathcal{S}_{m}}^{(r)}\: w_{dom}^{(r)}$. It is normalized such that $\sum\nolimits_{r} w_{\mathcal{S}_{m}}^{(r)}=1$. The global temporal feature for source data in domain $\mathcal{S}_{m}$ $\mathbf{t}_{\mathcal{S}_{m}}$ is therefore formulated as:
\begin{equation}
\label{eqn:method:glob-t}
    \mathbf{t}_{\mathcal{S}_{m}} = \sum\nolimits_{r}w_{\mathcal{S}_{m}}^{(r)} \: lt_{\,\mathcal{S}_{m}}^{(r)}.
\end{equation}
The global temporal feature for target data $\mathbf{t}_{\mathcal{T}}$ is defined similarly with the subscript $\mathcal{S}_{m}$ replaced by $\mathcal{T}$. However, as the target data are unlabeled, the weights of its temporal local features would depend solely on the \textit{dominance weight}, i.e., $w_{\mathcal{T}}^{(r)} = w_{dom}^{(r)}$, $\mathbf{t}_{\mathcal{T}} = \sum\nolimits_{r}w_{\mathcal{T}}^{(r)} \: lt_{\,\mathcal{T}}^{(r)}$.

With the global temporal features extracted, TAMAN aims to perform feature alignment for spatial and temporal features jointly. This is achieved by minimizing the moment-based feature discrepancies $d_{f}$ and $d_{\mathbf{t}}$ concurrently. Both $d_{f}$ and $d_{\mathbf{t}}$ are defined equivalently with Equation~\ref{eqn:method:dis-feat}. Overall, the objective function for TAMAN is expressed as:
\begin{equation}
\label{eqn:method:obj-taman}
    \begin{aligned}
    \mathcal{L}_{vms}
    &= \sum\limits_{j=1}^{M}\frac{1}{n_{\mathcal{S}_{j}}}\sum\nolimits_{v}
    L_{\mathcal{S}_{j}}(C_{\mathcal{S}_{j}}(\mathbf{t}_{v\,\mathcal{S}_{j}}), y_{v\,\mathcal{S}_{j}})\\
    &+ \lambda_{df}\:d_{f}(\mathcal{S},\mathcal{T})
    + \lambda_{d\mathbf{t}}\:d_{\mathbf{t}}(\mathcal{S},\mathcal{T}),
    \end{aligned}
\end{equation}
where $L_{\mathcal{S}_{j}}$ stands for the classification loss for the $j$-th source classifier with $y_{v\,\mathcal{S}_{j}}$ being the ground truth of the $v$-th input source video from domain $\mathcal{S}_{j}$, while $\lambda_{df}$ and $\lambda_{d\mathbf{t}}$ are the trade-off parameters for the moment-based spatial and temporal feature discrepancies, respectively.

During the testing phase, the target data are first propagated through the spatial and temporal feature extractors, and then the $M$ classifiers trained by source data. To obtain the final classification prediction, the outputs from each classifier $P_{j} = C_{\mathcal{S}_{j}} (\mathbf{t}_{\mathcal{T}}),\, j\in[1,M]$ are combined. The most intuitive method is to average all the outputs. Yet, since the domain shift between different source-target domain pairs are different, their target accuracies also vary. To address this issue, we propose a weighted ensemble schema to combine the outputs effectively. The idea behind the \textit{prediction weight} $w_{P}$ is that the final prediction should focus on the classifier whose output is of higher certainty. Given that the sum of the weights, i.e., $\sum\nolimits_{j=1}^{M}w_{P\,j}$ should be 1, the \textit{prediction weight} could be defined as:
\begin{equation}
\label{eqn:method:pred-w}
    w_{P\,j} = \sigma(\sum\limits_{c=1}^{\mathcal{C}} P_{j,\,c} \, log(P_{j, \,c})).
\end{equation}
Here $\mathcal{C}$ is the number of video classes, $P_{j,\,c}$ corresponds to the prediction of the $c$-th class from the $j$-th classifier, while $\sigma$ is the softmax function performed across the $M$ classifiers, i.e., $\sigma(x_{j}) = exp(x_{j})\,/\,\sum\nolimits_{j=1}^{M}exp(x_{j})$. The final prediction is therefore the weighted sum of predictions from each classifier guided by the \textit{prediction weight} $w_{P}$.

\begin{table*}[!ht]
\center
\resizebox{.9\linewidth}{!}{\noindent
\begin{tabular}{c|c|cccc|ccc}
\hline
\hline
  \multicolumn{2}{c|}{\multirow{2}{*}{Methods}} &
  \multicolumn{4}{c|}{\textbf{Daily-DA}} &
  \multicolumn{3}{c}{\textbf{Sports-DA}} \\
\cline{3-9}
\multicolumn{2}{c|}{} & \textbf{Daily}$\to$\textbf{A} & \textbf{Daily}$\to$\textbf{H} & \textbf{Daily}$\to$\textbf{M} & \textbf{Daily}$\to$\textbf{K} & \textbf{Sports}$\to$\textbf{U} & \textbf{Sports}$\to$\textbf{S} & \textbf{Sports}$\to$\textbf{K}\\
\hline
\parbox{0.12\linewidth}{\centering Source-only}
& TRN & 23.58$\pm$0.21 & 44.17$\pm$0.31 & 33.75$\pm$0.25 & 61.93$\pm$0.58 & 88.72$\pm$0.63 & 56.32$\pm$0.44 & 74.10$\pm$0.85 \\
\hline
\multirow{9}{*}{\parbox{0.12\linewidth}{\centering Adversarial-based}}
& s-DANN & 21.21$\pm$0.35 & 33.15$\pm$0.31 & 21.75$\pm$0.25 & 61.93$\pm$0.72 & 82.50$\pm$0.75 & 50.73$\pm$0.45 & 65.72$\pm$0.58 \\
& s-ADDA & 21.30$\pm$0.21 & 33.25$\pm$0.28 & 23.80$\pm$0.20 & 62.26$\pm$0.54 & 85.42$\pm$0.76 & 52.03$\pm$0.43 & 67.20$\pm$0.50 \\
& s-TA\textsuperscript{3}N & 21.76$\pm$0.16 & 39.91$\pm$0.39 & 33.75$\pm$0.30 & 61.75$\pm$0.55 & 83.76$\pm$0.66 & 53.52$\pm$0.56 & 73.15$\pm$0.75 \\
& c-DANN & 20.64$\pm$0.33 & 35.83$\pm$0.46 & 18.00$\pm$0.25 & 61.66$\pm$0.68 & 83.08$\pm$0.68 & 50.53$\pm$0.52 & 64.88$\pm$0.63 \\
& c-ADDA & 21.45$\pm$0.25 & 33.24$\pm$0.30 & 24.00$\pm$0.20 & 62.08$\pm$0.58 & 86.05$\pm$0.65 & 53.27$\pm$0.37 & 69.80$\pm$0.64 \\
& c-TA\textsuperscript{3}N & 22.24$\pm$0.20 & 40.42$\pm$0.32 & 33.80$\pm$0.30 & 62.18$\pm$0.60 & 84.68$\pm$0.72 & 55.76$\pm$0.48 & 74.30$\pm$0.87 \\
& MDAN & 23.35$\pm$0.38 & 43.33$\pm$0.42 & 33.00$\pm$0.40 & 61.52$\pm$0.42 & 87.96$\pm$0.82 & 57.04$\pm$0.46 & 72.86$\pm$0.54 \\
& DCTN & 24.84$\pm$0.36 & 44.14$\pm$0.48 & 34.25$\pm$0.30 & 62.16$\pm$0.44 & 88.84$\pm$0.64 & 57.36$\pm$0.38 & 67.58$\pm$0.60 \\
& MDDA & 22.73$\pm$0.26 & 45.30$\pm$0.35 & 35.00$\pm$0.50 & 63.21$\pm$0.63 & 89.81$\pm$0.70 & 57.63$\pm$0.40 & 74.48$\pm$0.66 \\
\hline
\multirow{7}{*}{\parbox{0.12\linewidth}{\centering Discrepancy-based}}
& s-MMD & 21.62$\pm$0.22 & 39.25$\pm$0.35 & 31.25$\pm$0.30 & 60.53$\pm$0.49 & 86.28$\pm$0.62 & 53.12$\pm$0.56 & 67.21$\pm$0.65 \\
& s-MCD & 23.80$\pm$0.28 & 39.95$\pm$0.36 & 32.00$\pm$0.25 & 61.63$\pm$0.57 & 87.36$\pm$0.62 & 57.08$\pm$0.43 & 74.50$\pm$0.75 \\
& s-CORAL & 21.51$\pm$0.15 & 38.76$\pm$0.26 & 33.00$\pm$0.25 & 61.35$\pm$0.45 & 86.10$\pm$0.52 & 53.72$\pm$0.32 & 68.75$\pm$0.45 \\
& c-MMD & 24.28$\pm$0.36 & 42.50$\pm$0.45 & 30.50$\pm$0.30 & 62.07$\pm$0.55 & 88.64$\pm$0.76 & 56.38$\pm$0.48 & 73.06$\pm$0.60 \\
& c-MCD & 25.68$\pm$0.28 & 44.45$\pm$0.33 & 33.50$\pm$0.25 & 62.92$\pm$0.78 & 90.22$\pm$0.80 & 58.47$\pm$0.21 & 74.54$\pm$0.62 \\
& c-CORAL & 23.96$\pm$0.16 & 41.78$\pm$0.24 & 34.25$\pm$0.25 & 61.38$\pm$0.33 & 87.36$\pm$0.58 & 57.98$\pm$0.32 & 74.36$\pm$0.44\\
& M3SDA & 24.83$\pm$0.23 & 42.50$\pm$0.35 & 33.25$\pm$0.50 & 62.21$\pm$0.41 & 88.75$\pm$0.70& 55.25$\pm$0.39 & 75.48$\pm$0.86 \\
\hline
Ours
& \textbf{TAMAN} & \textbf{29.95}$\pm$0.35 & \textbf{48.33}$\pm$0.38 & \textbf{36.75}$\pm$0.50 & \textbf{64.36}$\pm$0.69 & \textbf{92.26}$\pm$0.84 & \textbf{62.15}$\pm$0.52 & \textbf{79.12}$\pm$0.54 \\
\hline
\hline
\end{tabular}
}
\smallskip
\caption{Results for MSVDA on \textbf{Daily-DA} and \textbf{Sports-DA} datasets (mean $\pm$ std).}
\label{table:5-1-sota}
\end{table*}

\section{MSVDA Benchmarks}
\label{section:msvda-db}

There are very limited cross-domain benchmark datasets for VUDA and its variant tasks. For the few cross-domain datasets available such as UCF-HMDB\textsubscript{\textit{full}}~\cite{chen2019temporal} for standard VUDA and HMDB-ARID\textsubscript{\textit{partial}}~\cite{xu2021partial} for Partial Video Domain Adaptation (PVDA), the source domains are always constraint to be a single domain. To facilitate MSVDA research, we propose two sets of comprehensive benchmarks, namely the \textbf{Daily-DA} and the \textbf{Sports-DA} datasets. Both datasets cover extensive MSVDA scenarios and provide adequate baselines with distinct domain shifts to facilitate future MSVDA research. 

The \textbf{Daily-DA} dataset comprises of videos with common daily actions, such as drinking and walking. It is constructed from four action datasets: ARID (\textbf{A})~\cite{xu2021arid}, HMDB51 (\textbf{H})~\cite{kuehne2011hmdb}, Moments-in-Time (\textbf{M})~\cite{kay2017kinetics}, and Kinetics (\textbf{K})~\cite{monfort2019moments}. Among which, HMDB51, Moments-in-Time, and Kinetics are widely used for action recognition benchmarking collected from various public video platforms. ARID is a more recent dark dataset, comprised with videos shot under adverse illumination conditions. ARID is characterized by its low RGB mean value and standard deviation (std), which results in larger domain gap between ARID and other video domains. A total of 8 overlapping classes are collected, resulting in a total of 18,949 videos. When performing MSVDA, one dataset is selected as the target domain, with the remaining three datasets as the source domains, resulting in four MSVDA tasks: \textbf{Daily}$\to$\textbf{A}, \textbf{Daily}$\to$\textbf{H}, \textbf{Daily}$\to$\textbf{M}, and \textbf{Daily}$\to$\textbf{K}. The training and testing splits are separated following the official splits for each dataset.

The \textbf{Sports-DA} dataset contains videos with various sport actions, such as bike riding and rope climbing. It is built from three large-scale action datasets: UCF101 (\textbf{U})~\cite{soomro2012ucf101}, Sports-1M (\textbf{S})~\cite{karpathy2014large}, and Kinetics (\textbf{K}). Compared to \textbf{Daily-DA}, this dataset is much larger in terms of number of classes and videos. A total of 23 overlapping classes are collected, resulting in a total of 40,718 videos, making the \textbf{Sports-DA} dataset one of the largest cross-domain video datasets introduced. Its objective is to validate the effectiveness of MSVDA approaches on large-scale video data. Similar to the \textbf{Daily-DA} dataset, one dataset is selected as the target domain, with the remaining two datasets as the source domains when performing MSVDA, resulting in three MSVDA tasks: \textbf{Sports}$\to$\textbf{U}, \textbf{Sports}$\to$\textbf{S}, and \textbf{Sports}$\to$\textbf{K}. The training and testing splits are separated following the official splits.

\section{Experiments}
\label{section:exps}

In this section, we evaluate our proposed TAMAN by conducting cross-domain action recognition on MSVDA benchmarks proposed in Section~\ref{section:msvda-db}. We present superior results on both proposed benchmarks. Ablation studies and empirical analysis of TAMAN are also presented to justify our design.

\subsection{Experimental Settings}
\label{section:exps:settings}
Cross-domain action recognition tasks are performed on both the \textbf{Daily-DA} and \textbf{Sports-DA} datasets, with a total of 7 cross-domain settings as presented in Section~\ref{section:msvda-db}. Following standard UDA evaluation protocols~\cite{saenko2010adapting}, source videos are labeled while target videos are strictly unlabeled. All methods employ TRN~\cite{zhou2018temporal} as the feature extractor backbone, which is pretrained on ImageNet~\cite{deng2009imagenet}. All experiments are implemented with PyTorch~\cite{paszke2019pytorch} library. \textit{More detailed implementation specifications are provided in the Appendix}.

\subsection{Overall Results and Comparisons}
\label{section:exps:results}
We first compare TAMAN with various UDA/VUDA and MSDA approaches, which include: (i) adversarial-based methods: DANN~\cite{ganin2015unsupervised}, ADDA~\cite{tzeng2017adversarial}, TA\textsuperscript{3}N~\cite{chen2019temporal}, MDAN~\cite{zhao2018adversarial}, DCTN~\cite{xu2018deep} and MDDA~\cite{zhao2020multi}; and (ii) discrepancy-based methods: MMD~\cite{long2015learning}, MCD~\cite{saito2018maximum}, CORAL~\cite{sun2016return} and M3SDA~\cite{peng2019moment}. For UDA/VUDA approaches (i.e., DANN, ADDA, TA\textsuperscript{3}N, MMD, MCD, and CORAL), two strategies are employed: (i) single-best (`s-'), where the adaptation is performed for each source-target pair with the best result selected; and (ii) source-combined (`c-'), where all source domains are combined to form a domain. The results are presented in Table~\ref{table:5-1-sota}. Following~\cite{peng2019moment}, we report the mean and standard deviation (std) of the top-1 accuracy with 5 runs under identical network settings. For comparison, we also report the results of the backbone TRN trained with supervised source data only and tested on the target data.

Results in Table~\ref{table:5-1-sota} demonstrate the effectiveness of TAMAN, achieving the best results on all MSVDA tasks and outperforming all prior approaches by noticeable margins. Notably, TAMAN outperforms all image-based MSDA approaches (i.e., MSDA, MDAN, DCTN, and MDDA) consistently by the average of more than 10\% relative improvements in mean accuracy. This empirically justifies the effectiveness of constructing temporal attentive robust global temporal features which are more transferable while incorporating both spatial and temporal features for feature moment alignment. Further, it could be observed that for all prior UDA/VUDA and MSDA approaches, the adaptation results are inferior to that of the backbone TRN trained without any adaptation approaches in at least 1 MSVDA task. This suggests that all methods suffer from negative transfer.
In particular, the effect is more severe for \textbf{Daily-DA}, with an average of 12 out of 17 approaches evaluated suffering from the negative transfer. This owes to the fact that \textit{Daily-DA} dataset contains data collected from ARID with distinct statistical characteristics, resulting in larger cross-domain gaps.

\begin{table}[t]
\center
\resizebox{1.\linewidth}{!}{
\begin{tabular}{c|cc}
\hline
\hline
Methods & \textbf{Daily}$\to$\textbf{A} & \textbf{Daily}$\to$\textbf{H}\\
\hline
\textbf{TAMAN} & \textbf{29.95} & \textbf{48.33}\\
\hline
TAMAN w/o local confidence & 27.85 & 46.25\\
TAMAN w/o dominance & 28.32 & 45.42\\
TAMAN w/o local attention & 25.21 & 43.75\\
\hline
\hline
\end{tabular}
}
\caption{Ablation studies for \textit{local attention weight}.}
\label{table:5-2-local_attention}
\end{table}

\begin{table}[t]
\center
\resizebox{1.\linewidth}{!}{
\begin{tabular}{c|cc}
\hline
\hline
Methods & \textbf{Daily}$\to$\textbf{A} & \textbf{Daily}$\to$\textbf{H} \\
\hline
\textbf{TAMAN} & \textbf{29.95} & \textbf{48.33}\\
\hline
TAMAN w/o dominance & 28.32 & 45.42 \\
TAMAN w min. $d_{lt}^{(r)}$ dominance & 28.22 & 45.12\\
TAMAN w max. $d_{lt}^{(r)}$ dominance & 27.58 & 44.17\\
\hline
\hline
\end{tabular}
}
\caption{Ablation studies for obtaining \textit{dominance weight}.}
\label{table:5-3-dominance}
\end{table}

\begin{table}[t]
\center
\resizebox{1.\linewidth}{!}{
\begin{tabular}{c|cc}
\hline
\hline
Methods & \textbf{Daily}$\to$\textbf{A} & \textbf{Daily}$\to$\textbf{H}\\
\hline
\textbf{TAMAN} & \textbf{29.95} & \textbf{48.33}\\
\hline
TAMAN by avg & 29.02 & 47.58\\
TAMAN by src. only accuracy & 29.17 & 47.92\\
\hline
\hline
\end{tabular}
}
\caption{Ablation studies for prediction ensemble schemas.}
\label{table:5-4-ensemble}
\end{table}

\subsection{Ablation Studies}
\label{section:exps:ablation}
To further validate the efficacy of TAMAN and justify its design, we perform detailed ablation studies. The ablation studies are conducted from three perspectives: (i) \textit{local attention weight} and its components; (ii) different strategies for obtaining \textit{dominance weight}; and (iii) different prediction ensemble schemas. All ablation studies are conducted with the \textbf{Daily}$\to$\textbf{A} and \textbf{Daily}$\to$\textbf{H} tasks.

\textbf{\textit{Local attention weight}}. We evaluate TAMAN against 3 variants to justify the design of the \textit{local attention weight}: (a) \textbf{TAMAN w/o local confidence}, where the \textit{local confidence weights} are set to be equal for all local temporal features; (b) \textbf{TAMAN w/o dominance}, where the \textit{local attention weight} does not incorporate \textit{dominance weights}; and (c) \textbf{TAMAN w/o local attention}, where the global temporal features are built by additive aggregation of local temporal features. Results presented in Table~\ref{table:5-2-local_attention} clearly demonstrate the necessity of both the \textit{local confidence weight} and the \textit{dominance weight}, both of which help construct the robust global temporal features for alignment. By employing either weight, TAMAN learns more transferable temporal features given the better result compared to all approaches evaluated in Table~\ref{table:5-1-sota}. It is also noted that though \textbf{TAMAN w/o local attention} falls behind TAMAN by a notable margin, it still performs better than most image-based MSDA approaches, justifying the need for joint alignment of both spatial and temporal features.

\textbf{Obtaining \textit{dominance weights}}. We propose the \textit{dominance weight} which is obtained from the disparity between global and local feature discrepancies in Section~\ref{section:method:taman}. Alternatively, the \textit{dominance weight} could be obtained directly by comparing local temporal feature discrepancies. Therefore, we justify the current strategy for obtaining the \textit{dominance weight} by evaluating TAMAN against \textbf{TAMAN w/o dominance} and two other variants: (a) \textbf{TAMAN w min. $d_{lt}^{(r)}$ dominance}, where the global temporal features are set to focus on the local temporal feature with the minimum cross-domain moment discrepancy; and (b) \textbf{TAMAN w max. $d_{lt}^{(r)}$ dominance}, whose global temporal features attend to the local temporal feature with maximum cross-domain moment discrepancy. As shown in Table~\ref{table:5-3-dominance}, the results justify the design of the \textit{dominance weight} through the disparity of discrepancies. While the other two strategies are computed with ease, their inferior results to \textbf{TAMAN w/o dominance} show that the sub-optimal dominance weight may negatively affect the global temporal features.

\textbf{Prediction Ensemble Schemas}. TAMAN utilizes a weighted ensemble schema based on prediction certainty. To justify such an approach, we compare TAMAN with the following variants: (a) \textbf{TAMAN by avg}, whose final prediction is ensembled by directly averaging across outputs from each classifier; and (b) \textbf{TAMAN by src. only accuracy}, whose prediction is ensembled following~\cite{peng2019moment}, with the weights of each prediction output derived by the source only accuracy between each source-target domain pair. As demonstrated in Table~\ref{table:5-4-ensemble}, the performance improvement of the ensemble strategy in TAMAN is marginal, indicating that the domain-variant feature learning plays a more vital role in MSVDA. It is noted that though the ensemble method in~\cite{peng2019moment} is more effective than simple averaging, it requires the evaluation of source-only results with each individual source-target domain pair, resulting in more computation and less efficiency.

\section{Conclusion}
\label{section:concl}

In this work, we propose a novel method for tackling Multi-Source Video Domain Adaptation (MSVDA). In contrast to prior works where only spatial features are aligned, TAMAN deals with MSVDA by dynamically aligning both spatial and temporal feature moments. TAMAN also attends to dominant domain-invariant local temporal features with high local classification confidence and low disparity between global and local feature discrepancies. We further pioneer in introducing novel MSVDA benchmarks to facilitate future MSVDA research. Our proposed TAMAN tackles MSVDA well, supported by extensive experiments and ablation studies across the proposed MSVDA benchmarks.

\begin{table*}[!htbp]
\center
\smallskip\begin{tabular}{c|c|c|c}
\hline
\hline
ARID Class & HMDB51 Class & Moments-in-Time Class & Kinetics Class \\
\hline
Drink & drink & drinking & drinking shots \\
\hline
\multirow{3}{*}{Jump} & \multirow{3}{*}{jump} & \multirow{3}{*}{jumping} & jumping bicycle \\
                      &                       &                          & jumping into pool \\
                      &                       &                          & jumping jacks \\
\hline
Pick & pick & picking & picking fruit \\
\hline
Pour & pour & pouring & pouring beer \\
\hline
\multirow{4}{*}{Push} & \multirow{4}{*}{push} & \multirow{4}{*}{pushing} & pushing car \\
                      &                       &                          & pushing cart \\
                      &                       &                          & pushing wheelbarrow \\
                      &                       &                          & pushing wheelchair \\
\hline
Run & run & running & running on treadmill \\
\hline
\multirow{2}{*}{Walk} & \multirow{2}{*}{walk} & \multirow{2}{*}{walking} & walking the dog \\
                      &                       &                          & walking through snow \\
\hline
Wave & wave & waving & waving hand \\
\hline
\hline
\end{tabular}
\smallskip
\caption{List of overlapping classes collected for \textbf{Daily-DA} dataset.}
\label{table:s-1-daily}
\end{table*}

\begin{table*}[!htbp]
\center
\smallskip\begin{tabular}{c|c|c}
\hline
\hline
UCF101 Class & Sports-1M Class & Kinetics Class \\
\hline
Archery & archery & archery \\
\hline
\multirow{2}{*}{Baseball Pitch} & \multirow{2}{*}{baseball} & catching or throwing baseball \\
                                &                           & hitting baseball \\
\hline
\multirow{2}{*}{Basketball Shooting} & \multirow{2}{*}{basketball} & playing basketball \\
                                     &                             & shooting basketball \\
\hline
Biking & bicycle & riding a bike \\
\hline
Bowling & bowling & bowling \\
\hline
Breaststroke & breaststroke & swimming breast stroke \\
\hline
Diving & diving & springboard diving \\
\hline
Fencing & fencing & fencing (sport) \\
\hline
Field Hockey Penalty & field hockey & playing field hockey \\
\hline
Floor Gymnastics & floor (gymnastics) & gymnastics tumbling \\
\hline
\multirow{3}{*}{Golf Swing} & \multirow{3}{*}{golf} & golf chipping \\
                            &                       & golf driving \\
                            &                       & golf putting \\
\hline
Horse Race & horse racing & riding or walking with horse \\
\hline
Kayaking & kayaking & canoeing or kayaking \\
\hline
Rock Climbing Indoor & rock climbing & rock climbing \\
\hline
Rope Climbing & rope climbing & climbing a rope \\
\hline
Skate Boarding & skateboarding & skateboarding \\
\hline
\multirow{2}{*}{Skiing} & \multirow{2}{*}{skiing} & skiing crosscountry \\
                        &                         & skiing mono \\
\hline
Sumo Wrestling & sumo & wrestling \\
\hline
Surfing & surfing & surfing water \\
\hline
Tai Chi & t'ai chi ch'uan & tai chi \\
\hline
Tennis Swing & tennis & playing tennis \\
\hline
Trampoline Jumping & trampolining & bouncing on trampoline \\
\hline
Volleyball Spiking & volleyball & playing volleyball \\
\hline
\hline
\end{tabular}
\smallskip
\caption{List of overlapping classes collected for \textbf{Sports-DA} dataset.}
\label{table:s-2-sports}
\end{table*}

\begin{figure}[t]
\begin{center}
  \includegraphics[width=1.\linewidth]{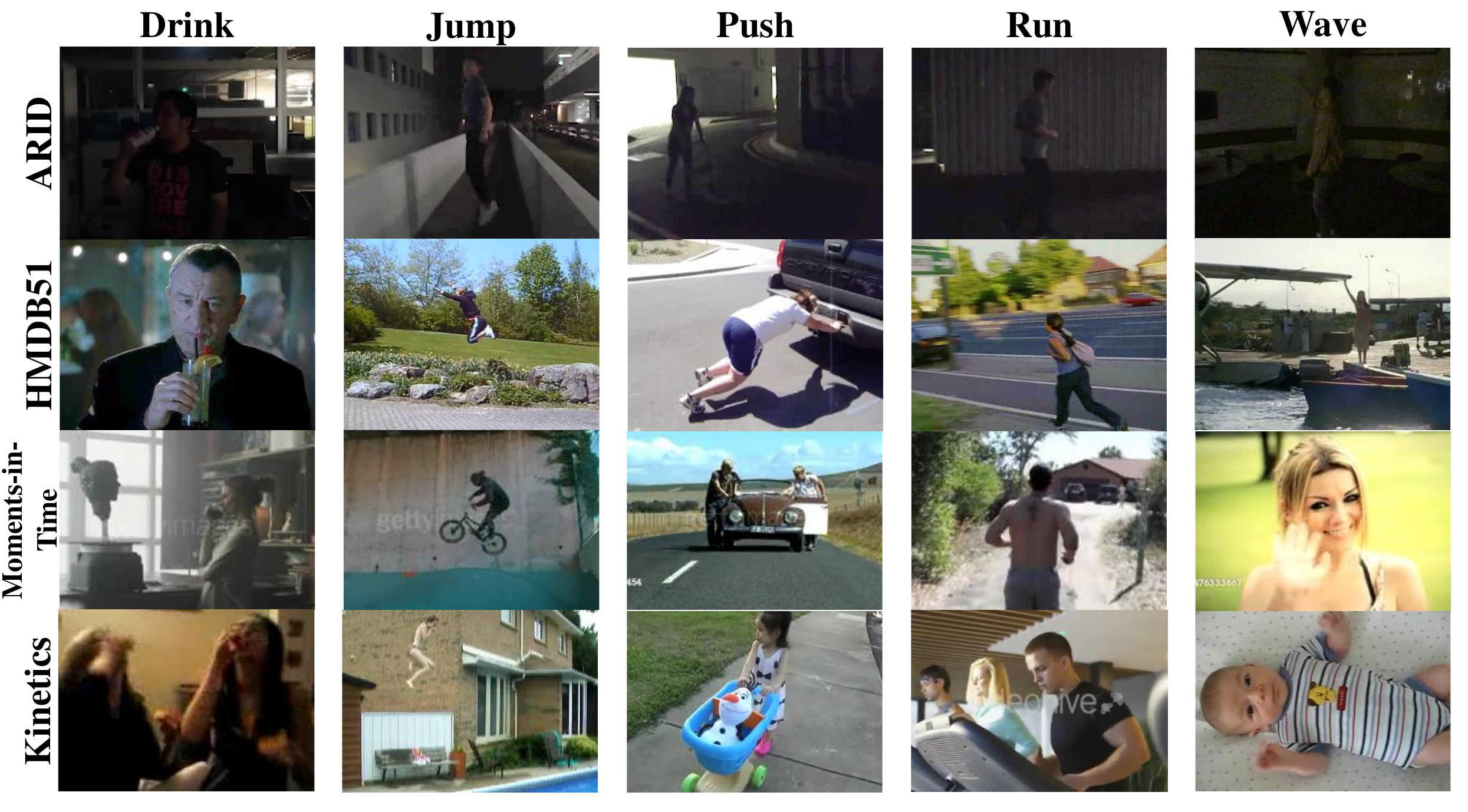}
\end{center}
  \caption{Sampled frames of videos from sampled classes in the \textbf{Daily-DA} dataset.}
\label{figure:s1-1-daily}
\end{figure}

\begin{figure}[t]
\begin{center}
  \includegraphics[width=1.\linewidth]{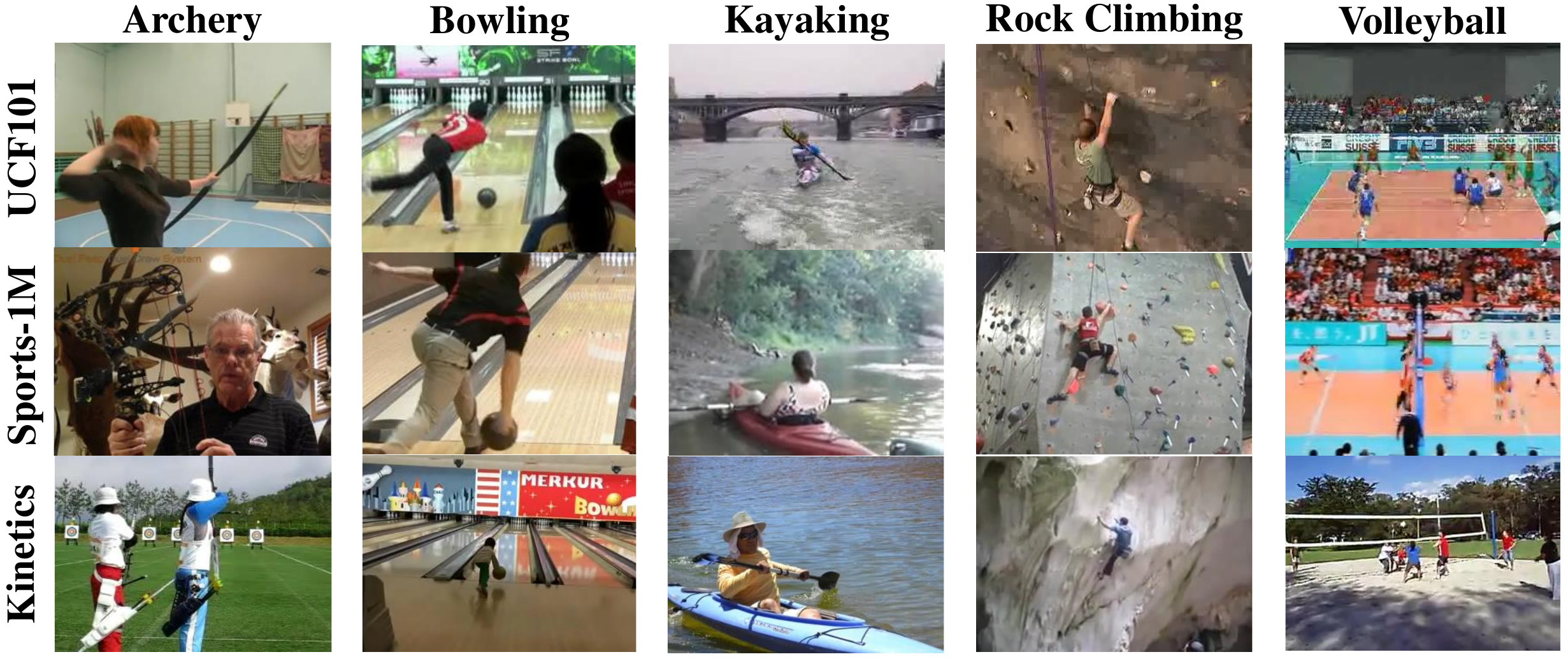}
\end{center}
  \caption{Sampled frames of videos from sampled classes in the \textbf{Sports-DA} dataset.}
\label{figure:s1-2-sports}
\end{figure}

\begin{figure*}[t]
\begin{center}
   \includegraphics[width=1.\linewidth]{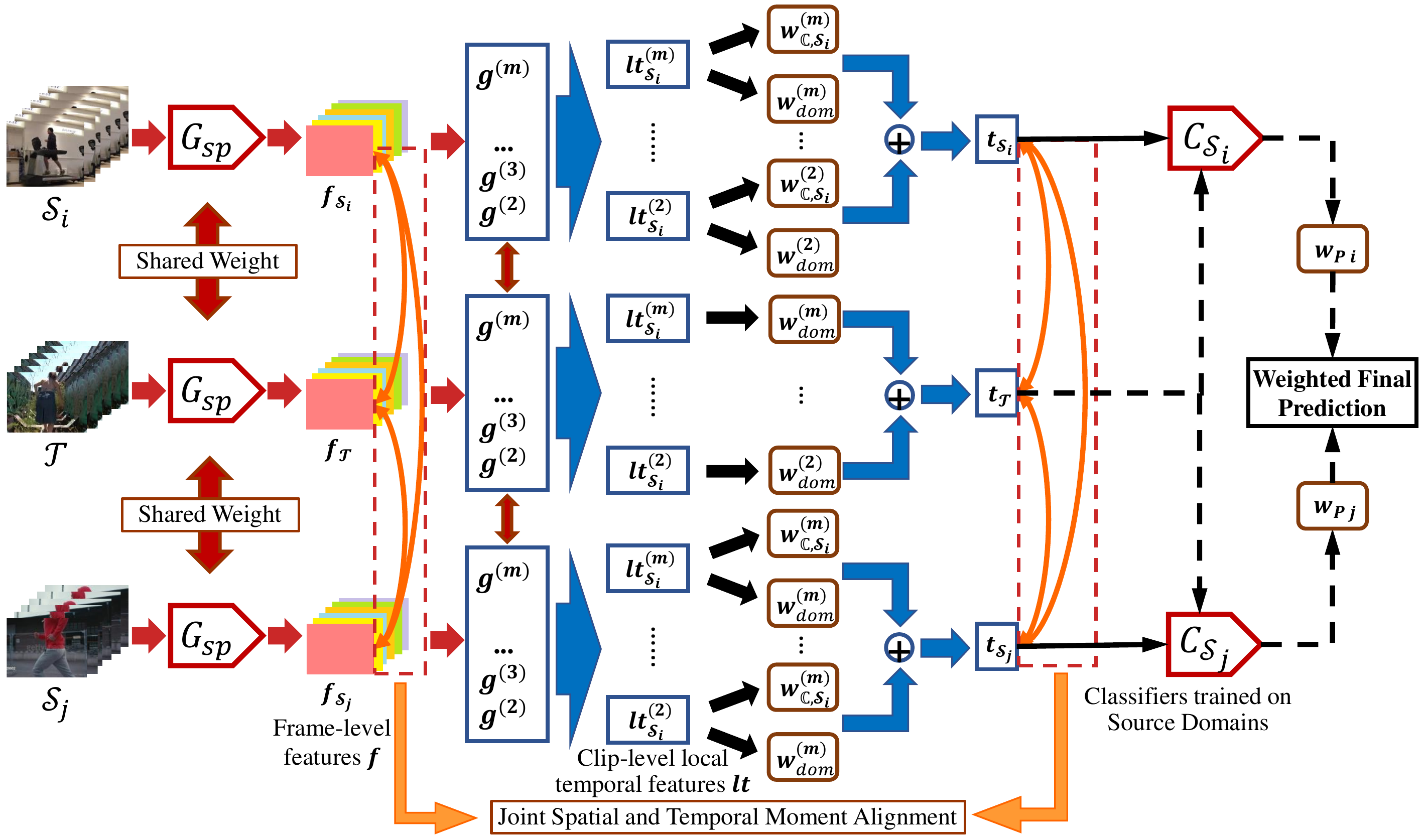}
\end{center}
    \smallskip\caption{Architecture of the proposed PATAN. \textit{Best viewed in color and zoomed in.}}
\label{figure:s2-3-patan}
\end{figure*}

\section{Appendix}
\label{section:supp}

\subsection{PVDA Benchmarks}
\label{section:supp:msvda-db}
In this work, we propose two sets of comprehensive benchmarks, namely the \textbf{Daily-DA} and the \textbf{Sports-DA} datasets, covering extensive \textit{Multi-Source Video Domain Adaptation} (MSVDA) scenarios to promote future MSVDA research. In this section, we describe each benchmark with more details.

\paragraph{\textbf{Daily-DA} dataset.}
The \textbf{Daily-DA} dataset comprises of videos with common daily actions. It is constructed from four action datasets: ARID (\textbf{A})~\cite{xu2021arid}, HMDB51 (\textbf{H})~\cite{kuehne2011hmdb}, Moments-in-Time (\textbf{M})~\cite{kay2017kinetics}, and Kinetics (\textbf{K})~\cite{monfort2019moments} (using the Kinetics-600 version). Among which, HMDB51, Moments-in-Time, and Kinetics datasets are widely used for action recognition benchmarking collected from various public video platforms (e.g.\ YouTube, Flickr). ARID is a more recent dataset, comprised with videos shot under adverse illumination conditions. Statistically, videos in ARID are characterized by their low RGB mean value and standard deviation, which results in larger domain gap between ARID and other video domains. A total of 8 overlapping classes are collected, which are listed in Table~\ref{table:s-1-daily}, resulting in a total of 18,949 videos. Among which, there are 2,776 training videos and 1,289 testing videos from ARID; 560 training videos and 240 testing videos from HMDB51; 4,000 training videos and 400 testing videos from Moments-in-Time; and 8,959 training videos and 725 testing videos from Kinetics. When performing MSVDA, one dataset is selected as the target domain, with the remaining three datasets as the source domains. We therefore construct four MSVDA tasks: \textbf{Daily}$\to$\textbf{A}, \textbf{Daily}$\to$\textbf{H}, \textbf{Daily}$\to$\textbf{M}, and \textbf{Daily}$\to$\textbf{K}. The training and testing splits are separated following the official splits for each dataset. Figure~\ref{figure:s1-1-daily} shows the comparison of sampled frames from sampled classes in the \textbf{Daily-DA} dataset.

\paragraph{\textbf{Sports-DA} dataset.}
The \textbf{Sports-DA} dataset comprises of videos with common sport actions, and is built from three large-scale action datasets: UCF101 (\textbf{U})~\cite{soomro2012ucf101}, Sports-1M (\textbf{S})~\cite{karpathy2014large}, and Kinetics (\textbf{K}) (also using the Kinetics-600 version). Compared to \textbf{Daily-DA}, this dataset is much larger in terms of both the number of classes and videos. A total of 23 overlapping classes are collected which are listed in Table~\ref{table:s-2-sports}, resulting in a total of 40,718 videos, making the \textbf{Sports-DA} dataset one of the largest cross-domain video datasets introduced. Among all the videos, there are 2,145 training videos and 851 testing videos from UCF101; 14,754 training videos and 1,900 testing videos from Sports-1M; and 19,104 training videos and 1,961 testing videos from Kinetics. As videos in both the original Sports-1M and Kinetics dataset are provided as YouTube links, we ensure that the collected videos are still valid. Invalid links are all omitted during collection. The \textbf{Sports-DA} dataset is designed to validate the effectiveness of MSVDA approaches on large-scale video data. Similar to the \textbf{Daily-DA} dataset, one dataset is selected as the target domain, with the remaining two datasets as the source domains when performing MSVDA, resulting in three MSVDA tasks: \textbf{Sports}$\to$\textbf{U}, \textbf{Sports}$\to$\textbf{S}, and \textbf{Sports}$\to$\textbf{K}. We follow the official split for separating the training and testing sets. Figure~\ref{figure:s1-2-sports} shows the comparison of sampled frames from sampled classes in the \textbf{Sports-DA} dataset.

\subsection{Detailed Implementation of Temporal Attentive Moment Alignment Network (TAMAN)}
\label{section:supp:detail-imp}
As presented in Section 3, we propose the Temporal Attentive Moment Alignment Network (TAMAN) to deal with the MSVDA task by constructing temporal attentive robust global temporal features while aligning spatial and temporal features jointly. The structure of our proposed TAMAN is depicted clearly in Figure~\ref{figure:s2-3-patan}. In this section, we further elaborate on the detailed implementation of TAMAN.

Our networks and all relevant experiments are implemented using the PyTorch~\cite{paszke2019pytorch} library. To obtain video features, we instantiate Temporal Relation Network~\cite{zhou2018temporal} with ResNet-101~\cite{he2016deep} as the backbone for video feature extraction for both source and target domain videos, with the model pretrained on ImageNet~\cite{deng2009imagenet}. The source and target feature extractors share parameters. New layers are trained from scratch, and their learning rates are set to 0.001. The pretrained layers which outputs the frame-level spatial features $f$ are frozen.

The stochastic gradient descent (SGD) algorithm~\cite{bottou2010large} is used for optimization, with the weight decay set to 0.0001 and the momentum to 0.9. The batch size is set to 128 per GPU. Our initial learning rate is set to 0.001 and is divided by 10 for three times during the training process. We train our networks with a total of 100 epochs for the \textbf{Daily-DA} dataset and a total of 40 epochs for the \textbf{Sports-DA} dataset. The trade-off weight for the moment-based spatial and temporal feature discrepancies $\lambda_{df}$ and $\lambda_{d\mathbf{t}}$ are set to 0.005 and 0.01. All experiments are conducted using two NVIDIA RTX 2080 Ti GPUs.

\clearpage
{\small
\bibliography{aaai}
}
\end{document}